\documentclass[twocolumn]{el-author}
\usepackage{xcolor}
\usepackage{graphicx}
\usepackage{epstopdf}

\date{12th December 2011}

\begin{document}

\title{Deep ensemble network with explicit complementary model for accuracy-balanced classification}

\author{Dohyun Kim, Kyeorye Lee, Jiyeon Kim, Junseok Kwon, and Joongheon Kim}

\abstract{
The average accuracy is one of major evaluation metrics for classification systems, while the accuracy deviation is another important performance metric used to evaluate various deep neural networks. 
In this paper, we present a new ensemble-like fast deep neural network, \emph{Harmony}, that can reduce the accuracy deviation among categories without degrading overall average accuracy.
Harmony consists of three sub-models, namely, Target model, Complementary model, and Conductor model.
In Harmony, an object is classified by using either Target model or Complementary model.
Target model is a conventional classification network for general categories, while Complementary model is a classification network especially for weak categories that are inaccurately classified by Target model.
Conductor model is used to select one of two models.  
Experimental results demonstrate that Harmony accurately classifies categories, while it reduces the accuracy deviation among the categories. 
}

\maketitle

\section{Introduction}
With increasing usage of deep learning in object classification problems~\cite{alexnet}, various deep neural networks have been proposed. 
To evaluate these deep neural networks, the average accuracy is widely used as a major performance metric. 
However, high average accuracy does not necessarily indicate a good classifier, as pointed out earlier in many studies~\cite{imbalance}. 
For example, large accuracy deviations indicate that the model has weaknesses for classifying certain objects.
In practice, the lower bound of accuracy becomes more important than the average accuracy because the degradation of service experience, due to the weaknesses for particular objects, can be more lethal than the gains from strengths. 
Therefore, it is desirable to have the accuracy deviation as small as possible while maintaining the original average accuracy.
One of the major factors that causes the accuracy deviation is the class imbalance problem that frequently occurs when the training data of particular classes (as called minor classes) is less than the learning data of other classes (as called major classes).
This class imbalance problem can be effectively solved by re-sampling based methods that generate meaningful data artificially and balance categories in the training dataset~\cite{SMOTE}.
However, although the dataset is balanced, the accuracy deviation problem can still occur. 

Given the sufficient dataset, the fundamental solution is to enhance the model structure toward increasing the classification accuracy and simultaneously reducing accuracy deviation.  
However, improving the model structure or developing novel models are time-consuming tasks, which is difficult to apply in practice. 
Therefore, it is required to reduce the accuracy deviation without changing the structure of the existing model. 
For example, re-sampling based methods increase training data for classes that are not accurately classified by using data generation techniques. 
The aforementioned approach is efficient when the dataset is not sufficient to generalise minor classes.
However, if the dataset is sufficient, this approach can make noises in the dataset and can degrade the overall classification performance.
To reduce the accuracy deviation, another approach multiplies weights to the loss function differently for each classes.
However, this approach also has the same problem as re-sampling based methods, in which the performance for weak classes are improved but the performance degradation occurs heavily in strong classes due to over-fitting.
Several ensemble-based methods such as bagging and boosting~\cite{imbalance} solve the aforementioned problem by using multiple models and demonstrate their robustness. 
However, conventional ensemble methods have limitations that a large number of models are required to gain striking performance improvement, which significantly increases computation time.

In this paper, we tackle the aforementioned issue and focus on reducing the accuracy deviation for each category in classification problems with balanced dataset. 
We solve performance deviation problems without performance degradation and present a novel deep neural network \textit{Harmony}, which consists of Target model, Complementary model, and Conductor model.
Target model is a classification network for general categories, while Complementary model is for weak categories that are wrongly classified by Target model.
Conductor model determines which classification model is used.

\section{Deep ensemble network with Explicit Complementary Model}

\begin{figure}[t]
    \centering
    \includegraphics[scale=0.27]{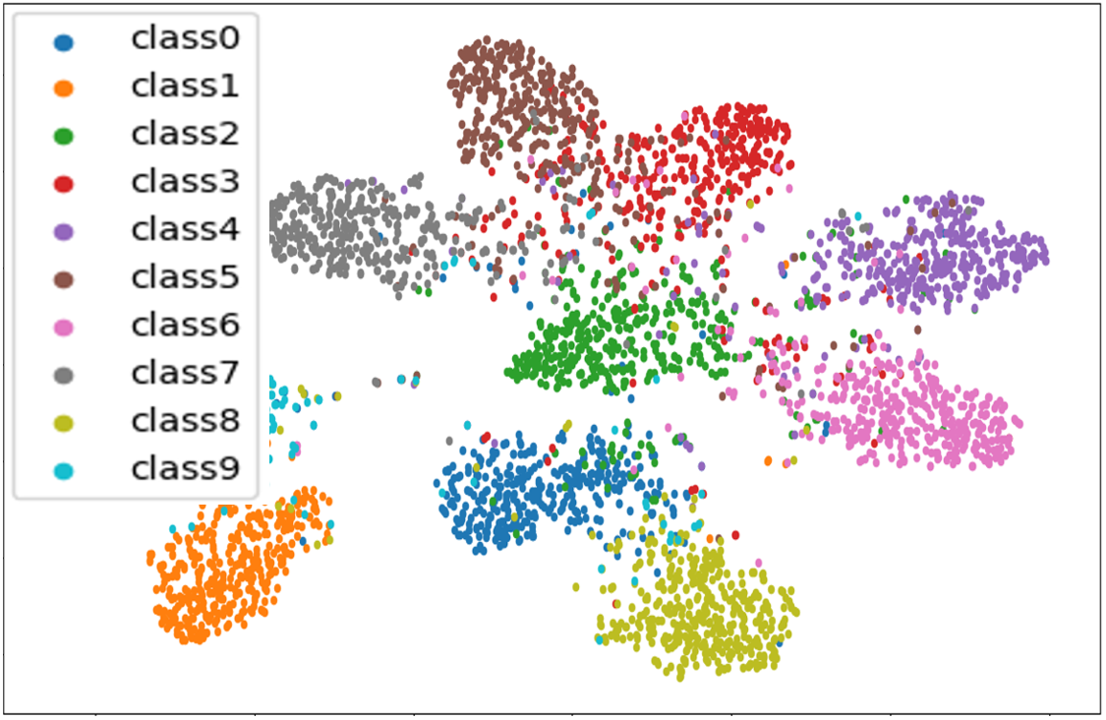}
    \caption{The embedded result for each class of Target model only, based on the neural network model, GoogLeNet.}
    	\label{fig:result(a)}
\end{figure}

The weak data that cause the accuracy deviation is typically inter-correlated and miss-classified. 
Fig.~\ref{fig:result(a)} shows classification results on the cifar10 dataset, which are embedded by using T-SNE~\cite{tsne}.
Table.~\ref{tab:1}(b) reports the accuracy for each class.
For this experiment, we employ GoogLeNet.
In Fig.~\ref{fig:result(a)}, we can observe that results of classes $2$, $3$, and $5$ are correlated to each other. 
In particular, classes $3$ and $5$ correspond to cat and dog, respectively, which is known to be one of the difficult subjects to distinguish each other~\cite{dogandcat}.
Our first intuition is that complementary model for the weak classes is needed to break down this correlation and accurately classify these correlated weak classes. 
The concept of the complementary model is also introduced in~\cite{mcl}, in which their goal is to improve overall performance.
Unlike~\cite{mcl}, we aim to reduce accuracy deviation without performance degradation, thus we have to fit the weak classes more strongly to complementary model. 
This over-fitted model works as noises that cause overall degradation in traditional majority ensemble based approach~\cite{mcl}.
In order to solve this problem, a large number of models are used, which increases the computation time in proportion to the number of models.
Our second intuition is that conductor model can pre-classify whether input data belongs to strong classes or weak classes and can determine which models (i.e. Target model and Complementary model) would be advantageous to process. 
In summary, our network consists of three modules, i.e. target model, complementary model and conductor model. 

\begin{figure}[t]
    \centering
    \includegraphics[scale=0.4]{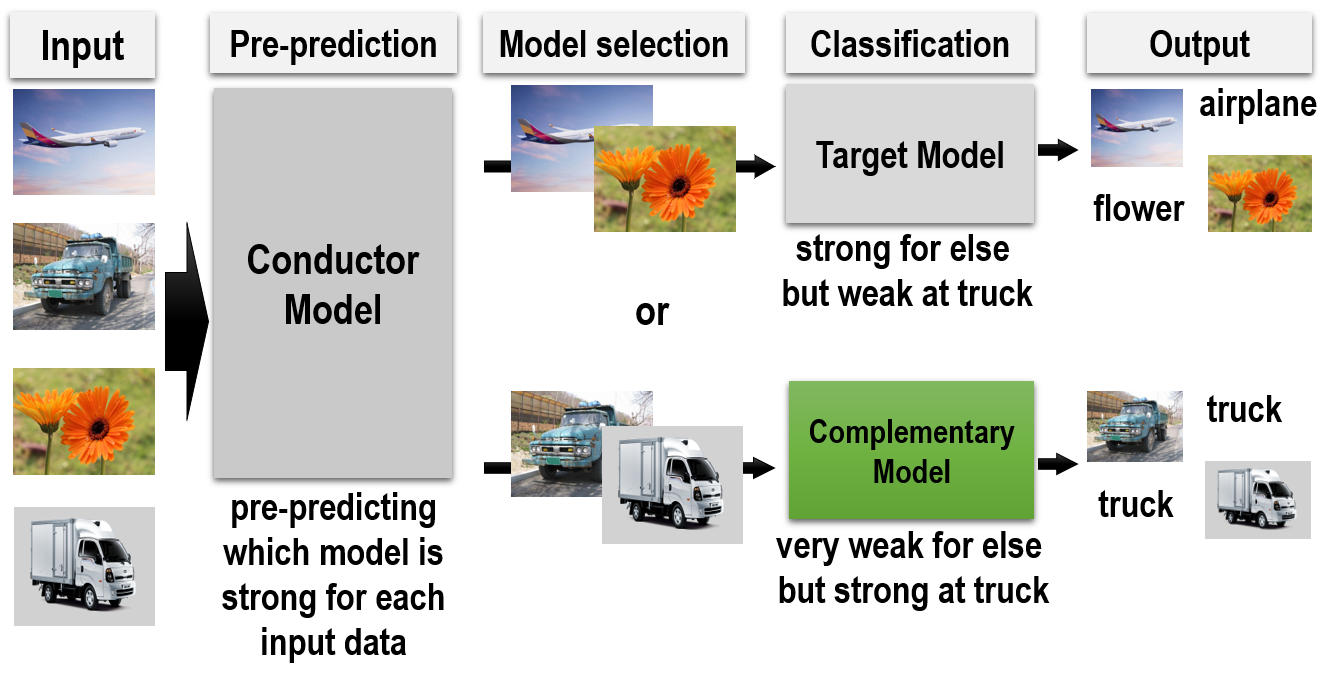}
    \caption{\emph{Harmony} consists of three networks, i.e., \emph{Target model}, \emph{Complementary model} and \emph{Conductor model}.}
    	\label{fig:HarmonyPipeline}
\end{figure}
Fig.~\ref{fig:HarmonyPipeline} shows the entire pipeline of the proposed algorithm. 
The description of each model is as follows.

\begin{itemize} 
    \item {\bf{Target model}} is a conventional classification network. It is weak for some classes (e.g. classes $2$, $3$ , and $5$ in Fig.~\ref{fig:result(a)}) and strong for else. 

    \item {\bf{Complementary model}} is a classification network that is intentionally trained to have a complementary performance distribution for target model with biased training dataset.
    Complementary model tries to accurate classify the classes that is wrongly classified by the target model.
    Note that the complementary model adopts exactly the same neural network as the target model. 
    Complementary model may over-fit the weak classes of target model, which induces the overall performance degradation. 
    However, this problem can be mitigated by following conductor model.

    \item {\bf{Conductor model}} determines which of Target model Complementary model would be advantageous to classify before the actual classification. 
    The classes are grouped into either weak or strong classes by binary classification of Conductor model.
    Note that Conductor model also adopts the similar neural network as Target and Complementary models, while the number of nodes in the last output layer is adjusted to two. 
\end{itemize}  

With three models, we follow divide and conquer approaches by dividing difficult problems into easy sub-problems. 
As shown in embedding results of Fig.~\ref{fig:result(a)}, classes $2$, $3$, and $5$ are hard to distinguish from each other because they are inter-correlated.
In this case, it is much simpler to train Conductor model to differentiate classes $2$, $3$, and $5$ from other classes.
And it is also relatively easy to make Complementary model to classify only weak classes. 

In Fig.~\ref{fig:result(a)}, because weak classes form one cluster, we can solve the accuracy deviation problem with a single Complementary model. 
However, if there are more groups that are difficult to classify, we need many Complementary models. 
In this case, Conductor model becomes a $1+|\mathcal{C}|$-nary classifier, and the time complexity still approximates $\mathcal{O}(2\mathcal{N})$ as in case of $|\mathcal{C}| = 1$, where $\mathcal{C}$ stands for the set of Complementary models and $\mathcal{N}$ stand for the time required to process one piece of data through the neural network, respectively.

\section{Evaluation}

\begin{table*}[t]
\caption{Numerical results}
  \centering
    \begin{tabular}{r||c|c|c|c|c|c|c||c|c|c||c||c}
    \hline\hline
    \multicolumn{1}{l||}{}       & \multicolumn{7}{c||}{Strong classes}                           & \multicolumn{3}{c||}{Weak classes} & Avg & 
    
    Var \\ \hline\hline
    class      & 0        & 1          & 4     & 6     & 7     & 8     & 9     & 2          & 3          & 5         &         &          \\ \cline{2-13} 
                                 & airplane & automobile & deer  & frog  & horse & ship  & truck & bird       & cat        & dog       &         &          \\ \hline\hline
    target (a)  & 0.901    & 0.953      & 0.867 & 0.924 & 0.922 & 0.94  & 0.932 & 0.827      & 0.719      & 0.835     & 0.8820   & 0.00466 \\ \hline
    target\_weighted (b)   & 0.848    & 0.95       & 0.861 & 0.864 & 0.892 & 0.925 & 0.907 & 0.815      & 0.82       & 0.801     & 0.8683  & 0.00246  \\ \hline
    complemantary (c)      & 0.813    & 0.926      & 0.795 & 0.819 & 0.793 & 0.92  & 0.891 & 0.878      & 0.852      & 0.864     & 0.8551  & 0.00240\\ \hline
    conductor (d) & 0.942    & 0.986      & 0.878 & 0.855 & 0.892 & 0.986 & 0.989 & 0.857      & 0.886      & 0.921     & 0.9192  & 0.00287 \\ \cline{2-13} 
                                 & \multicolumn{7}{c||}{0.932}                                    & \multicolumn{3}{c||}{0.888}          &         &          \\ \hline
    harmony (e) & 0.88     & 0.952      & 0.851 & 0.885 & 0.883 & 0.938 & 0.926 & 0.854      & 0.832      & 0.852     & 0.8853  & 0.00150  \\ \hline
    bagging (n = 2) (f)          & 0.868    & 0.931      & 0.82  & 0.883 & 0.892 & 0.93  & 0.912 & 0.784      & 0.801      & 0.815     & 0.8636  & 0.00270 \\ \hline
    bagging (n = 5) (g)          & 0.869    & 0.934      & 0.83  & 0.886 & 0.9   & 0.927 & 0.913 & 0.796      & 0.822      & 0.822     & 0.8699  & 0.00220 \\ \hline
    target + target2 (h)         & 0.907    & 0.965      & 0.884 & 0.925 & 0.929 & 0.923 & 0.928 & 0.83       & 0.755      & 0.838     & 0.8884  & 0.00397  \\ \hline\hline
    \end{tabular}
  \label{tab:1}
\end{table*}

\vspace{-3mm}

Table.\ref{tab:1} shows numerical results based on GoogLeNet for each model with image classification dataset, Cifar10, consisting of $50,000$ training and $10000$ test images for $10$ categories.
Table.~\ref{tab:1}(a) shows results of Target model in which the accuracy deviation problem can occur even if the dataset is balanced. 
Table.~\ref{tab:1}(b) are results of the model trained by increasing loss weights for the weak classes on the same model as in (a).

\begin{figure}[t]
    \centering
    \includegraphics[width = 0.96\linewidth]{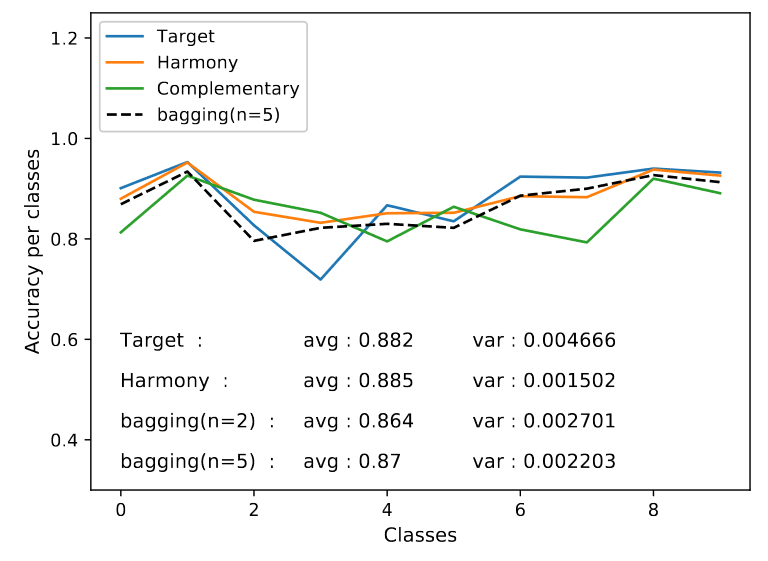}
    \caption{Performance evaluation results with image classification dataset, Cifar10, based on the GoogLeNet.}
    	\label{fig:result(b)}
\end{figure}

The model increased the classification accuracy for the weak classes, but the overall performance degraded due to the over-fitting. 
Table.~\ref{tab:1}(c) are results of Complementary model trained more biased than (b). 
Table.~\ref{tab:1}(d) shows the accuracy of Conductor model that groups weak and strong classes of Target model by binary classification. 
The accuracy  means the ratio of correctly classified data to total number of data for each class. 
Table.~\ref{tab:1}(e), (f), and (g) show the accuracy for each class, average accuracy and variance, when using the proposed harmony method, bagging with two weak models and five weak models respectively.
The weak model indicates a model with low performance and relatively low average accuracy. 
Fig.~\ref{fig:result(b)} visualises them. 
As demonstrated in Fig.~\ref{fig:result(b)}, although the accuracy deviation is reduced to some extent, bagging with two weak models yielded the low performance. 
In case of bagging with five weak models, the performance deviation is further reduce. 
As the number of models used increases, the overall performance will be improved, but the processing time will also be increased accordingly. 
On the other hand, Harmony is more effective in all cases in terms of accuracy deviation. 
In particular, it reduced the variance of accuracy by $ 0.0046 $ to $ 0.0015 $ by $ 68 \% $ compared to Target model without any performance degradation, even it was rather improved a little. 
Harmony is less effective in terms of improving the average performance compared to (h) which are results of ensemble two different target models. 
However, the main purpose of (h) is to improve the average accuracy not to reduce accuracy deviation, it is less effective in terms of accuracy deviation, even it becomes worse. 

\vspace{-1mm}

\section{Conclusion and discussion}
We aimed to solve the problem of accuracy deviation without performance degradation when the dataset is balanced. To handle the problem that the processing time increases in proportion to the number of models used in the conventional majority based ensemble, we propose a managed ensemble approach \textit{harmony} with explicit complementary model.  In experiments, we used only one Complementary model because there is one weak class. 
But for more complex problems, \textit{harmony} can show good performance even with the increase in the number of models. 
We have experimented only on the image classification task, but we expected \textit{harmony} can be widely applied to other domains with its generality.

\vskip3pt
\ack{This work was supported by National Research Foundation of Korea (2016R1C1B1015406).}

\vskip5pt

\noindent D. Kim, K. Lee, J. Kim, J. Kwon, and J. Kim (\textit{Chung-Ang Univ.})
\vskip3pt

\noindent E-mails: jskwon@cau.ac.kr ,  joongheon@cau.ac.kr (Corresponding authors: J. Kwon and J. Kim)

\end{document}